\documentclass[journal,comsoc]{IEEEtran}

\usepackage{cite}

\usepackage{amsmath,amssymb,amsfonts}
\usepackage{algorithmic}
\usepackage{graphicx}
\usepackage{textcomp}
\usepackage{xcolor}
\usepackage{float}
\usepackage[nolist]{acronym}
\usepackage[none]{hyphenat}

\usepackage[shortlabels]{enumitem}
\usepackage{tabularx}
\usepackage{comment}
\usepackage[normalem]{ulem}
\usepackage{url}

\usepackage{cancel}
\usepackage{placeins}
\usepackage{cite}

\usepackage{gensymb}
\usepackage[misc,geometry]{ifsym} 
\usepackage{makecell}
\usepackage{wrapfig}

\usepackage{breakurl}

\def\BibTeX{{\rm B\kern-.05em{\sc i\kern-.025em b}\kern-.08em
    T\kern-.1667em\lower.7ex\hbox{E}\kern-.125emX}}

\newif\ifrev
\revfalse 
\ifrev
\newcommand{\revmaj}[1]{\textcolor{blue}{#1}}
\else
\newcommand{\revmaj}[1]{#1}
\fi

\begin{document}

\title{User-Level Handover Decision Making Based on Machine Learning Approaches}

\author{João~P.~S.~H.~Lima,~Alvaro~A.~M.~de~Medeiros,~Eduardo~P.~de~Aguiar,~Vicente~A.~de~Sousa~Jr.~and~Tarciana~C.~B.~Guerra
\thanks{João Lima is with CPQD (e-mail: jsales@cpqd.com.br). Alvaro Medeiros and Eduardo Aguiar are with Federal University of Juiz de Fora, Brazil (e-mails: \{alvaro, eduardo.aguiar\}@engenharia.ufjf.br). Alvaro Medeiros is also with Munster Technological University, Cork, Ireland. Vicente Sousa and Tarciana Guerra are with Federal University of Rio Grande do Norte, Brazil (e-mails: \{tarciana.guerra.051, vicente.sousa\}@ufrn.edu.br). This study was financed in part by FUNTTEL/Finep and the Coordena\c{c}\~{a}o de Aperfei\c{c}oamento de Pessoal de N\'{i}vel Superior - Brasil (CAPES) - Finance Code 001. The proof of concept simulations provided by this Letter was supported by High Performance Computing Center~(NPAD/UFRN).} \thanks{Digital Object Identifier: 10.14209/jcis.2022.11}
}


\maketitle

\begin{acronym}
  \acro{1G}{First Generation}
  \acro{2G}{Second Generation}
  \acro{3G}{Third Generation}
  \acro{4G}{Fourth Generation}
  \acro{5G}{Fifth Generation}
  \acro{5GRA}{Remote Areas applications}
  \acro{5GPHY}{5G Physical Layer}
  \acro{ADC}{analogic-to-digital converter}
  \acro{AGC}{automatic gain control}
  \acro{ASIP}{Application Specific Integrated Processors}
  \acro{AWGN}{additive white Gaussian noise}
  \acro{BDTM}{burst data transfer mode}
  \acro{BER}{bit error rate}
  \acro{BS}{base station}
  \acro{CDTM}{continuous data transfer mode}
  \acro{CFO}{Carrier Frequency Offset}
  \acro{CHF}{Characteristic Function}  
  \acro{CoMP} {Cooperative Multi-point}
  \acro{CP}{cyclic prefix}
  \acro{CR}{Cognitive Radio}
  \acro{CS}{cyclic suffix}
  \acro{CSI}{channel state information}
  \acro{CSMA}{carrier sense multiple access}
  \acro{DFT}{discrete Fourier transform}
  \acro{DPD}{digital pre-distortion}
  \acro{DZT}{discrete Zak transform}
  \acro{eMBB}{enhanced mobile broadband}
  \acro{EPC}{evolved packet core}
  \acro{FBMC}{Filter-bank multi-carrier}
  \acro{FDE}{frequency-domain equalizer}
  \acro{FDMA}{frequency division multiple access}
  \acro{FD-OQAM-GFDM}{frequency-domain OQAM-GFDM}
  \acro{FEC}{forward error control}
  \acro{FPGA}{Field Programmable Gate Array}
  \acro{FTN}{Faster than Nyquist}
  \acro{FT}{Fourier transform}
  \acro{FSC}{frequency-selective channel}
  \acro{GFDM}{Generalized Frequency Division Multiplexing}
  \acro{GS-GFDM}{guard-symbol GFDM}
  \acro{HPA}{high power amplifier}
  \acro{IBI}{inter-block interference}  
  \acro{ICI}{inter-carrier interference}
  \acro{IDFT}{Inverse Discrete Fourier Transform}
  \acro{IFI}{inter-frame interference}
  \acro{IMS}{IP multimedia subsystem}
  \acro{IoT}{Internet of Things}
  \acro{IP}{Internet Protocol}
  \acro{IQ}{in-phase and quadrature}
  \acro{ISI}{inter-symbol interference}
  \acro{IUI}{inter-user interference}
  \acro{KPI}{key performance indicator}
  \acro{LDPC}{low density check parity code}
  \acro{LLR}{log-likelihood ratio}
  \acro{LMMSE}{linear minimum mean square error}
  \acro{LTE}{Long-Term Evolution}
  \acro{LTE-A}{Long-Term Evolution - Advanced}
  \acro{M2M}{Machine-to-Machine}
  \acro{MA}{multiple access}
  \acro{MAC}{medium access control layer}
  \acro{MF}{Matched filter}
  \acro{MIMO}{multiple-input multiple-output}
  \acro{MMSE}{minimum mean square error}
  \acro{MRC}{maximum ratio combiner}
  \acro{MSE}{mean-squared error}
  \acro{mMTC}{massive machine type communication}
  \acro{MTC}{machine type communication}
  \acro{MU}{multi user}
  \acro{NEF}{noise enhancement factor}
  \acro{NFV}{network functions virtualization}
  \acro{OFDM}{Orthogonal Frequency Division Multiplexing}
  \acro{OOB}{out-of-band}
  \acro{OQAM}{Offset Quadrature Amplitude Modulation}
  \acro{PAPR}{peak to average power ratio}
  \acro{PHY}{physical layer}
  \acro{PRBS}{Pseudo Random Bit Sequence}
  \acro{PSD}{Power Spectrum Density}
  \acro{QAM}{quadrature amplitude modulation}
  \acro{QPSK}{quadrature phase shift keying}
  \acro{QoE}{Quality of Experience}
  \acro{QoS}{Quality of Service}
  \acro{RC}{raised-cosine}
  \acro{RF}{radio frequency}
  \acro{ROF}{roll-off factor}
  \acro{RRC}{root raised cosine}
  \acro{SC}{single carrier}
  \acro{SC-FDE}{Single Carrier Frequency Domain Equalization}
  \acro{SC-FDMA}{Single Carrier Frequency Domain Multiple Access}
  \acro{SCD}{Successive Cancellation Decoding}
  \acro{SDN}{software-defined network}
  \acro{SDR}{software-defined radio}
  \acro{SDW}{software-defined waveform}
  \acro{SEP}{symbol error probability}
  \acro{SER}{symbol error rate}
  \acro{SIC}{successive interference cancellation}
  \acro{SISO}{single-input single-output}
  \acro{SMS}{Short Message Service}
  \acro{SNR}{signal-to-noise ratio}
  \acro{ST}{space-time}
  \acro{STO}{Symbol Timing Offset}
  \acro{STC}{space time code}
  \acro{STFT}{short-time Fourier transform}
  \acro{TD-OQAM-GFDM}{time-domain OQAM-GFDM}
  \acro{TR-STC}{time-reversal space-time coding}
  \acro{TR-STC-GFDMA}{TR-STC Generalized Frequency Division Multiple Access}
  \acro{TVC}{time-variant channel}
  \acro{TVWS}{TV white space}
  \acro{UHF}{Ultra High Frequency}
  \acro{URLL}{ultra-reliable low latency}
  \acro{V2V}{vehicle-to-vehicle}
  \acro{VHF}{Very High Frequency}
  \acro{V-OFDM}{Vector OFDM}
  \acro{ZF}{zero-forcing}
  \acro{W-GFDM}{windowed GFDM}
  \acro{WHT}{Walsh-Hadamard Transform}
  \acro{WLAN}{wireless Local Area Network}
  \acro{WLE}{widely linear equalizer}
  \acro{WLP}{wide linear processing}
  \acro{WRAN}{Wireless Regional Area Network}
  \acro{WSN}{wireless sensor networks}
  \acro{TLS}{transport layer security}
  \acro{BRICS}{Brazil, Russia, India, and South Africa}
  \acro{LPWAN}{low-power wide-area network}
  \acro{UE}{user equipment}
  \acro{GPS}{Global Positioning System}
  \acro{RFID}{Radio-frequency identification}
  \acro{CAN1}{Controller Area Network 1}
  \acro{CAN2}{Controller Area Network 2}
  \acro{P2P}{Peer-to-peer}
  \acro{ICIC}{inter-cell interference coordination}
  \acro{ACK}{acknowledgment}
\end{acronym}

\begin{abstract}

This Letter covers a broad comparison of methods for classification and regression applications for a user-level handover decision making in scenarios with adverse propagation conditions involving buildings, coverage holes, and shadowing effects. The simulation campaigns are based on network simulator~\textit{ns-3}. The comparison encompasses classical machine learning approaches, such as KNN, SVM, and neural networks, but also state-of-the-art fuzzy logic systems and latter boosting machines. The results indicate that SVM and MLP are the most suitable for the classification of the best handover target, although fuzzy system SOFL can perform similarly with lower processing time. Additionally, for the download time estimation, LightGBM provides the smallest error with short processing time, even in hard propagation scenarios. 
\end{abstract}


\section{Introduction}
\label{intro}

Two key concepts adopted by next-generation networks are cell densification 
and operation at high frequencies. 
\revmaj{Although larger bandwidth is available, enabling higher data rates, the propagation on higher frequencies limits the cell coverage area.}


\revmaj{A fundamental cellular procedure} directly affected by this scenario is the handover \revmaj{(HO)}, which is the transfer of a communication session (e.g., a call, a video stream, a file download) from one cell to another without loss or interruption of service. 
Since the \revmaj{User Equipment (UE)} must switch between physical channels during such procedure, it requires very rapid decisions from the cellular network in order to guarantee the \revmaj{Quality of Experience (QoE)}. 
The number of handovers is expected to increase notably, specially considering propagation-intensive scenarios (e.g. \revmaj{high-frequency urban cells}, outdoor-to-indoor coverage) whose severe propagation situations cause areas with meaningful signal degradation, creating non-deterministic coverage holes.


Three characteristics are required from evolved handover procedures in order to provide solid work in upcoming mobile communication systems: \textbf{seamless} (no interruption); \textbf{spectral-efficiency aware} (controlled signaling load); and \textbf{smart} (decision-making leveraged by machine learning and the vast amount of information available in the network).


The current \revmaj{HO} schemes in 3GPP networks (4G and 5G) are set upon some characteristic events, as depicted in Table~\ref{tab:eventos}~\cite{3gpp}. \revmaj{UEs} are supposed to provide frequent measurement reports to \revmaj{the base station, known as enhanced Node-B (eNB)} in 4G standard, containing received signal metrics, such as Reference Signal Received Power (RSRP) and Reference Signal Received Quality (RSRQ). 
Accordingly, the \revmaj{HO} procedures occur with simple power level comparisons based on events of those measurements, being denominated as deterministic \revmaj{HO}s. Despite its simple implementation, this configuration may lead to numerous inefficient, unnecessary or ping-pong \revmaj{HO}s, flooding network channels with counterproductive signaling load, and degrading spectral efficiency.


\begin{table}[t]
    \centering
    \caption{Characteristic events for 3GPP Handover~\protect\cite{3gpp}.}
    \label{tab:eventos}
    \begin{tabular}{ll}
    \hline\noalign{\smallskip}
        Event & \makecell{Description }\\
        \noalign{\smallskip}\hline\noalign{\smallskip}
        A1 & \makecell{Primary cell (PC) signal power becomes better than a threshold} \\
        
        A2 & \makecell{PC signal power becomes worse than a threshold}  \\
        
        A3 & \makecell{Secondary cell (SC) signal power becomes better than PC \\ by an offset} \\
        
        A4 & \makecell{SC signal power becomes better than a threshold}  \\
    \noalign{\smallskip}\hline    
    \end{tabular}
\end{table}


Machine learning \revmaj{(ML)} applications to develop smarter handovers are numerous. The authors in \cite{bang2019bayesian} implement a Bayesian regression method for \revmaj{HO} improvements in high-speed trains in South Korea, whereas authors in~\cite{yan2019} applies \emph{K-Nearest Neighbours} \revmaj{(KNN)} for possible real-time \revmaj{HO} decisions in vehicular networks. 
In~\cite{zhohov2021latency}, ML approaches are used to reduce latency and classify the best cell available for \revmaj{HO}. In~\cite{saeed2017fuzzy} and \cite{saeed2018qlearning} fuzzy logic and reinforcement learning is used to for optimize traditional \revmaj{HO} parameters, such as time-to-trigger and \revmaj{HO} margin. The work from~\cite{elmahdy2021tuning} compares different computational intelligence models for \revmaj{HO} parameter tuning while solution in~\cite{xia2019traffic} employs~\emph{LightGBM} to predict mobile network traffic. A lane-changing algorithm for autonomous vehicles is developed in~\cite{gu2019xgboost} based on~\emph{Extreme Gradient Boosting}. The authors of~\cite{tayyab} promote a rich survey on \revmaj{HO} management \revmaj{and} \revmaj{4G and 5G} tendencies, while~\cite{stamou2019} offers a vast survey on autonomous \revmaj{HO} management in the heterogeneous network context. In~\cite{tarcianapaper}, the development of neural networks in \revmaj{HO} mechanism were implemented at different levels, and an extensive database was produced. These works have demonstrated the capacity of different configurations of neural networks-handover integration to outperform classical 3GPP \revmaj{HO} methods.

As a plenty of methodologies are developed for smart \revmaj{HO}, this work aims to bring a broad comparison of methods in user-level scenarios of mobility in 3GPP networks, based in data set from~\cite{tarcianapaper}. In this Letter, 
the classification addresses the determination of the best \revmaj{HO} target, whereas the regression estimates the time and percentage of download that a mobile user performs while moving and requesting a \revmaj{HO}. Coverage holes and shadowing effects are modeled into simulation scenarios to emulate an urban environment. 
The coverage hole is a complete lack of coverage in a particular region, representing a connection interruption, e.g., due to a mmWave severe propagation condition. 
Thus, this Letter extends the solution and results of~\cite{tarcianapaper} with the following contributions:

\begin{itemize}
    \item Enhancing the comparison of \revmaj{ML classification approaches}, 
    including \revmaj{most recent} fuzzy logic systems;
    \item Estimating download duration and percentage of completion, providing new inputs for \revmaj{HO} decision, using classical and state-of-the-art approaches, such as latter boosting machines (not increasing data acquisition complexity to feed learning algorithms);
    \item Evaluating of fuzzy-based \revmaj{HO} methods on urban scenarios with the presence of buildings, shadowing effects, and coverage holes (evaluation also includes network-related \revmaj{Key Performance Indicators (KPIs)} and algorithms' processing time). 
\end{itemize}




\section{System Modelling and Evaluated Scenarios}
\label{sec:systemmodel}

\begin{figure}[t]
    \centering
    \includegraphics[width=0.29\textwidth]{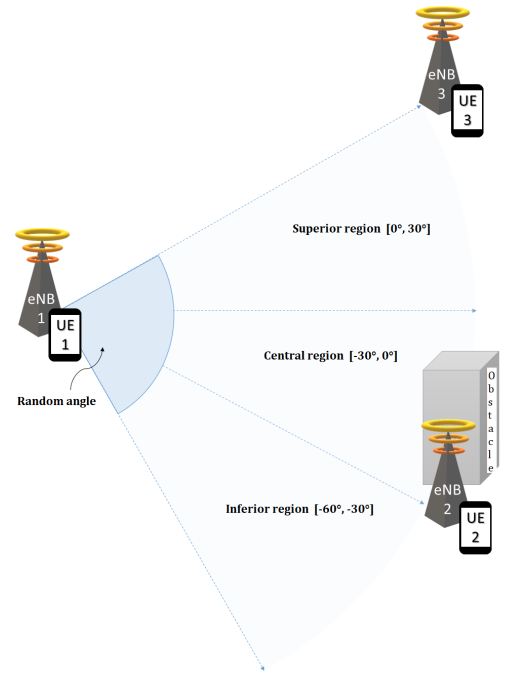}
    \caption{The simulation environment.}
    \label{fig:simulation_config}
\end{figure}

The environment setup relies on~\cite{ali2015} using version 3.22 of Network Simulator~\emph{ns-3}, as presented in Fig.~\ref{fig:simulation_config}. It counts with~3 eNBs,~3 UEs, and an obstacle near eNB~2.


The UE~1 starts simultaneously to download a file and move straight with a random angle from~$-60$\degree~ to~$30$\degree, with a constant speed of~60~km/h. Quickly, 
\revmaj{it} escapes from the coverage area of eNB~1 and enters the coverage area of eNBs~2 and~3, requesting handover. 
For each scenario, about~1200 runs are analyzed. The levels of RSRP and RSRQ are captured every~200~ms, \revmaj{feeding} the input database \revmaj{to evaluate ML} algorithms. The download process uses \revmaj{the well-known} TCP protocol 
with the file size of~15~MB. 
Finally, the simulation is carried for~100~s. 


In this \revmaj{Letter}, two scenarios are explored. The first one is based on the Okumura-Hata propagation model, pondering only path loss as large scale attenuation, chosen for being a widely used deterministic model for characterizing urban and suburban areas. This scenario may represent a situation with averaged RSRQ and RSRP, in which instantaneous values are filtered (e.g., moving average filter), flattening shadowing and small-scale fading effects. The second scenario sums the random shadowing effect to the Okumura-Hata model, indicating measurements that are more resembling to the fluctuations of RSRP and RSRQ. For both scenarios, the coverage hole is modeled by the presence of a building whose dimensions are extensive enough to emulate a region of connection interruption, with a very high path loss~\cite{ali2015}. 

%
%
%
%
%
    
    
    
    
    
    
    
    
    
    


More details about scenario modeling, including simulation parameters and the SINR Radio Environment Maps (REMs) of eNB~2 for both scenarios can be found in~\cite{tarcianapaper}.



\section{The Proposed Evaluation}
\label{sec:methods}


The authors of \cite{tarcianapaper} develop studies to explore the possibilities of how the machine learning models can be integrated with the handover decision making process coordinated by the eNBs, but they did not include most recent gradient boosting machines nor fuzzy systems. Moreover, in this work a statistical analysis is also developed to corroborate the initial results presented. Nonetheless, this 
work also implements new metric predictions, which are regression machine learning tasks. The prediction of a download time and percentage of completion can be of high value for network architects in order to design optimal handover algorithms. A handover triggering can be considered unnecessary if a solid estimation indicates that the ongoing download will be completed. On the other hand, a handover shall be anticipated if the estimation indicates that it will not be completed.




Thus, similar to \cite{tarcianapaper}, this Letter searches for the best eNB in terms of download completion and duration (the classification problem), including state-of-the-art fuzzy systems. The authors of~\cite{angelovempirical} and~\cite{angelovautonomous} have indicated that fuzzy strategies are capable of providing equivalent (or even better) performances while consuming less computational resources 
\revmaj{ compared to traditional Artificial Intelligence}  tools. They had never been applied to the \revmaj{HO} problem targeted by this Letter. 
We also propose regression techniques to estimate the percentage of completed download and the download duration. 


The methods applied for the \revmaj{HO} decision (classification problem) are \revmaj{Autonomous Learning Multimodel System (ALMMo)~\cite{angelovautonomous} (which has no tuning parameters, since it extracts all the features and adjustments from data); Self-Organizing Fuzzy Logic Classifier (SOFL)~\cite{gu} (using \textit{Mahalanobis} distance and \textit{Granularity Level} of 2.9); Type-2 Fuzzy Logic Classifier (T2FLS)~\cite{calderano,deaguiar} (being the learning parameter $\alpha = 0.01$, tolerance $\epsilon=10{^-8}$, $\beta_{1}=0.9$ and $\beta_{2}=0.999$); Support Vector Machine Classifier (SVM)~\cite{elmahdy2021tuning,aghabozorgi2019} (with linear kernel and penalty parameter of 10 and 100 for Scenarios 1 and 2, respectively); and Multilayer Perceptron Classifier (MLP)~\cite{haykin,campos} (with 6 neurons in hidden layers and solver \textit{lbfgs}).}

Regarding the regression problem (estimation of the completed download percentage and the download duration), six methods are compared: \revmaj{Multilayer Perceptron Regressor (MLP)~\cite{haykin,campos} (with 22 and 4 neurons in hidden layers, \textit{tanh} and \textit{logistic} activation functions and \textit{lbfgs} solver for Scenarios 1 and 2, respectively); \revmaj{KNN}~\cite{yan2019,haykin} with 4 and 6 neighbors considered for each Scenario; Random Forest (RF)~\cite{elmahdy2021tuning,zhang2019} with 94 and 106 trees in the forest, in each case; Gradient Boosting Machine (GBM)~\cite{nesa2018}, with 84 and 120 trees in their ensemble; Extreme Gradient Boosting (XGBoost)~\cite{gu2019xgboost}, with 174 and 120 estimators each; and Light Gradient Boosting Machine (LightGBM)~\cite{xia2019traffic}, which used 148 and 139 estimators in the ensemble for each Scenario.}

There are some considerations on how the \revmaj{ML} models could be implemented in a real network. First, it would be necessary a setup step, in which the network would act without the models' action because it is necessary to collect/store data and to train models. In our analysis, the RSRP and RSRQ  measurements occur in this initial phase, along with the information about download completions and their required times. They are models' inputs so that they can be trained. This also clarifies how important is the processing time
, since the models must be updated as fast as possible, with no harm to the network. After this initial phase, the models would be available to act on the \revmaj{HO} decision making at the eNBs. Furthermore, if 
\revmaj{network alters} e.g., an introduction of a new eNB, it would be necessary to retrain the models to account \revmaj{such changes}~\cite{tarcianapaper}.

\section{Results and Discussion}
\label{sec:results}

The selected methods were tested with Python~3.7 and Matlab scripts on an i7-6700HQ processor computer (2.6~GHz). 
To \revmaj{provide} statistical robustness, \revmaj{the~\emph{k-Fold} technique~\cite{haykin} was implemented}, with $k$~=~5. \revmaj{Additionally}, the test was carried~33 times for each method employed. 
In order to facilitate the reproducibility of the proposal discussed in this Letter, all the codes \revmaj{for training, testing, and the resulting parameters of all algorithms, besides our simulation campaign} numerical data are available \revmaj{in~\cite{githubnotebooks}}.

\subsection{Handover Decision (Classification Problem)}
For the classification of the best eNB for \revmaj{HO}, three metrics were used for comparison: the average prediction accuracy (the percentage of correct predictions by the classifier), its standard deviation and processing time. \revmaj{The initial results are presented in Tables~\ref{tab:initial_noshadow} and~\ref{tab:initial_shadow}, for Scenarios~1 
and~2, respectively.}

\begin{table}[t]
    \centering
    \caption{Initial results for Scenario 1.}
    \label{tab:initial_noshadow}
    \begin{tabular}{llll}
    \hline
        Method & Accuracy (\%) & Std. Dev. & Time (s)\\\hline
        MLP & 99.72 & 0.10 & 5.25 \\
       
        SVM & 99.74 & 0.08 & 0.54 \\
        
        SOFL & 99.11 & 0.22 & 12.51 \\
        
        T2FLS & 98.94 & 0.68 & 5845.78 \\
        
        ALMMo & 99.61 & 0.11 & 257.47 \\
    \noalign{\smallskip}\hline
    \end{tabular}
\end{table}

\begin{table}[t]
    \centering
    \caption{Initial results for Scenario 2.}
    \label{tab:initial_shadow}
    \begin{tabular}{llll}
    \hline\noalign{\smallskip}
        Method & Accuracy (\%) & Std. Dev. & Time (s)\\
    \noalign{\smallskip}\hline\noalign{\smallskip}
        MLP & 86.22 & 0.96 & 15.94 \\
        
        SVM & 86.34 & 0.33 & 21.26 \\
        
        SOFL & 85.32 & 0.71 & 14.60 \\
        
        T2FLS & 72.40 & 0.98 & 7672.62 \\
        
        ALMMo & 68.06 & 1.16 & 357.74 \\
    \noalign{\smallskip}\hline
    \end{tabular}
\end{table}

\revmaj{We verify the statistical validity of the data obtained from the proposed solution by using the two-sample $t$-test~\cite{moore},}
%
\revmaj{whose $t$ parameter is given by
\begin{equation} \label{t-test}
 t = \frac{{{{\bar G}_1} - {{\bar G}_2}}}{{\sqrt {\frac{{{s_{{G_1}}}^2}}{l} + \frac{{{s_{{G_2}}}^2}}{h}} }},
\end{equation}
where ${\bar G}_1$ and ${\bar G}_2$ are the means, $s_{{G_1}}$ and $s_{{G_2}}$ the standard deviation and~$h$ and~$l$ are the size of samples~$G_{1}$ and~$G_{2}$, respectively.
In addition to the evaluation of~$t$, it is also important to infer the hypothesis $H_0: {{\bar G}_1} = {{\bar G}_2}$ and $H_1={{\bar G}_1} \ne {{\bar G}_2}$,}
%
\revmaj{where the null hypothesis $H_{0}$ indicates that both~${G}_1$ and~${G}_2$ methods have obtained the same accuracy, while~$H_{1}$ is the alternative hypothesis which indicates that the accuracy levels are distinct.
Given a significance level $\alpha_t$, the $p$-value, which is calculated from~$t$-test, represents the lowest possible value to reject~$H_{0}$~\cite{moore}. Values lower than~$\alpha_t$ indicates the rejection of~$H_{0}$ in $(1-\alpha_t)\times100\%$ of the cases }\revmaj{(i.e., if $p$-value $< \alpha_t$, the alternative hypothesis $H_1$ is valid)}. 
Here, we consider~$\alpha_t$~=~0.05. 

In this stage, the $t$-test \revmaj{compares} the performance of the \revmaj{adopted} strategies for obtaining the greatest accuracy on classifying the best \revmaj{HO} target for UE~1. \revmaj{Table~\ref{tab:ttest_noshadow} and~\ref{tab:ttest_shadow} present the evaluations for Scenario~1 and~2, respectively. In both Tables, the $p$-value is presented, as well as the confidence interval on the difference of the population means, and the hypothesis inferred (0 for $H_0$ and 1 for $H_1$). }


\begin{table}[t]
    \centering
    \caption{$T$-test results for Scenario 1.}
    \label{tab:ttest_noshadow}
    \begin{tabular}{llllll}
    \hline\noalign{\smallskip}
        ${G}_1$ & ${G}_2$ & $p$-value & Low. b. & Upp. b.  & \revmaj{${H}$} \\
        \noalign{\smallskip}\hline\noalign{\smallskip}
        SVM  & ALM & 3.56E-07 & 0.0009 & 0.0018 & 1 \\
        
        \textbf{SVM} & \textbf{MLP} & \textbf{0.425} & \textbf{-2.70E-4} & \textbf{6.34E-4} & \textbf{0}\\
        
        SVM & SOFL & 4.66E-19 & 0.0055 & 0.0071 & 1 \\
        
        SVM & T2FL & 1.17E-07 & 0.0056 & 0.0105 & 1 \\
        
        MLP & ALM & 2.13E-05 & 0.0007 & 0.0017 & 1 \\
        
        MLP & SOFL & 5.12E-19 & 0.0053 & 0.007 & 1 \\
        
        MLP & T2FL & 1.83E-07 & 0.0054 & 0.0103 & 1 \\
        
        SOFL & ALM & 9.99E-16 & -0.0058 & -0.0041 & 1 \\
        
        \textbf{SOFL} & \textbf{T2FL} & \textbf{0.1803} & \textbf{-0.0008} & \textbf{0.0042} & \textbf{0} \\
        
        ALM & T2FL & 3.39E-06 & 0.0042 & 0.0091 & 1 \\
    \noalign{\smallskip}\hline
    \end{tabular}
\end{table}

\begin{table}[t]
    \centering
    \caption{$T$-test results for Scenario 2.}
    \label{tab:ttest_shadow}
    \begin{tabular}{llllll}
    \hline\noalign{\smallskip}
        ${G}_1$ & ${G}_2$ & $p$-value & Low. b. & Upp. b.  & \revmaj{${H}$} \\
        \noalign{\smallskip}\hline\noalign{\smallskip}
        SVM  & ALM & 1.26E-44 & 0.1786 & 0.1871 & 1 \\
        
        \textbf{SVM} & \textbf{MLP} & \textbf{0.5019} & \textbf{-0.0024} & \textbf{0.0048} & \textbf{0} \\
        
        SVM & SOFL & 1.66E-9 & 0.0075 & 0.013 & 1 \\
        
        SVM & T2FL & 2.47E-44 & 0.1358 & 0.1431 & 1 \\
        
        MLP & ALM & 1.86E-60 & 0.1764 & 0.1869 & 1 \\
        
        MLP & SOFL & 5.76E-5 & 0.0049 & 0.0132 & 1 \\
        
        MLP & T2FL & 7.31E-57 & 0.1334 & 0.143 & 1 \\
        
        SOFL & ALM & 6.14E-55 & 0.1679 & 0.1773 & 1 \\
        
        SOFL & T2FL & 1.78E-54 & 0.125 & 0.1334 & 1 \\
        
        ALM & T2FL & 2.47E-24 & -0.0487 & -0.0381 & 1 \\
        \noalign{\smallskip}\hline
    \end{tabular}
\end{table}

Based on Tables~\ref{tab:initial_noshadow} and~\ref{tab:ttest_noshadow}, the results indicate that all methods perform optimally when there are no shadowing effects to disturb predictions. However,~\emph{SVM} and \emph{MLP} have the best scores and reduced processing time. The $p$-value of the $t$-test 
\revmaj{is greater than~$\alpha_t$} only when comparing \emph{SVM} to \emph{MLP} and \emph{SOFL} to \emph{T2FL}, which indicates the validity of the null hypothesis ($H = 0$). Therefore, the $t$-test demonstrates there is no statistical difference between these algorithms in these cases and it confirms that the best models for this classification task are \emph{SVM} and \emph{MLP}. Moreover, the fuzzy logic-based~\emph{ALMMo} also demonstrates excellent accuracy, but it fails to deliver it quickly. \revmaj{We credit the longer time required for T2FLS and ALMMo mainly due to the training process. ALMMo extracts features autonomously, without further parameters and form its structure empirically from the observed data. The T2FLS on the other hand requires larger pre-processing calculations that could affect the training phase.}

Furthermore, looking at the results for Scenario~2 on Tables~\ref{tab:initial_shadow} and~\ref{tab:ttest_shadow}, in which the shadowing effects are present, some algorithms are still achieving reasonable precision, especially~\emph{SVM} and~\emph{MLP} classifiers, although the accuracy falls considerably (around~13\%). Again, they outperform the others on the comparison, and they do not present relevant differences to each other, confirmed by the $t$-test. However, in this case, it is important to accentuate that fuzzy rule-based~\emph{SOFL} classifier was capable of reaching competitive accuracy while requiring the shortest processing time, which is meaningful to the fuzzy logic context.

\subsection{Download Time Estimation (Regression Problem)}
For regression, each Scenario demanded a different regression variable. For the first one, the prediction was made for the download duration, since the majority of the downloads are able to be completed within simulation time. However, for the second one, the prediction was made for the percentage of completed download, seeing that it is a more challenging scenario and the majority of downloads could not be completed at the end of 100~s of simulation.
Therefore, the analyzed metrics are the mean absolute error ($MAE$) between the actual download time and the predicted value, the standard deviation, and processing time. The results are presented in Tables~\ref{tab:reg_results_noshadow}~and~\ref{tab:reg_results_shadow} below, for Scenarios~1 and~2, respectively.

\begin{table}[t]
    \centering
    \caption{Regression results for Scenario 1.}
    \label{tab:reg_results_noshadow}
    \begin{tabular}{llll}
    \hline\noalign{\smallskip}
        Method & $MAE$ & Std. Dev. & Time (s)\\
        \noalign{\smallskip}\hline\noalign{\smallskip}
        MLP & 0.27163 & 0.00800 & 21.85 \\
        
        KNN & 0.12791 & 0.00490 & 1.24 \\
        
        RF & 0.11987 & 0.00524 & 49.34 \\
        
        GBM & 0.12452 & 0.00727 & 32.98 \\
        
        XGBoost & 0.12343 & 0.00714 & 25.67 \\
        
        LightGBM & 0.11553 & 0.00317 & 14.81 \\
    \noalign{\smallskip}\hline
    \end{tabular}
\end{table}

\begin{table}[t]
    \centering
    \caption{Regression results for Scenario 2.}
    \label{tab:reg_results_shadow}
    \begin{tabular}{llll}
    \hline\noalign{\smallskip}
        Method & $MAE$ & Std. Dev. & Time (s)\\
        \noalign{\smallskip}\hline\noalign{\smallskip}
        MLP & 6.32415 & 0.07108 & 8.98 \\
        
        KNN & 6.00046 & 0.07525 & 1.68 \\
        
        RF & 5.99726 & 0.05373 & 53.05 \\
       
        GBM & 6.19593 & 0.07841 & 34.15 \\
        
        XGBoost & 5.88485 & 0.10150 & 17.53 \\
        
        LightGBM & 5.08845 & 0.04832 & 13.76 \\
        \noalign{\smallskip}\hline
    \end{tabular}
\end{table}

We notice on Table~\ref{tab:reg_results_noshadow}  that all methods presented a relatively low $MAE$, being~\emph{LightGBM} the most accurate with $MAE = 0.11553$; and \emph{MLP} being the least accurate with $MAE = 0.27163$. In Table~\ref{tab:reg_results_shadow}, due to the presence of shadowing, the values of mean absolute error are greater, as expected. Again, the most precise was~\emph{LightGBM} 
and the least precise was~\emph{MLP}.

Regarding execution time,~\emph{KNN} obtained the best result, possibly explained by the database not being so numerous and the hyperparameter $K$ being considerably small, reducing the computational cost. The second lowest time was presented by~\emph{LightGBM}, which has processing speed as an advantage. Differently,~\emph{Random Forest} was the slowest, probably due to the number of trees created during training.

The two-sample $t$-test~\cite{moore} is applied once again, now seeking statistical differences between regression methods $G_1$ and $G_2$. Hence, Tables~\ref{tab:ttest_reg_noshadow} and~\ref{tab:ttest_reg_shadow} are obtained for the first and second Scenarios, respectively. 

\begin{table}[t]
    \centering
    \caption{$T$-test results for regression in Scenario 1.}
    \label{tab:ttest_reg_noshadow}
    \begin{tabular}{llllll}
    \hline\noalign{\smallskip}
        ${G}_1$ & ${G}_2$ & $p$-value & Low. b. & Upp. b.  & \revmaj{${H}$} \\
        \noalign{\smallskip}\hline\noalign{\smallskip}
        XGB & Light & 6.85E-7 & 0.0052 & 0.0106 & 1\\
        
        \textbf{XGB} & \textbf{GBM} & \textbf{0.5785} & \textbf{-0.0047} & \textbf{0.0027} & \textbf{0} \\
        
        XGB & MLP & 8.11E-40 & -0.1530 & -0.1406 & 1 \\
        
        XGB & RF & 0.0493 & -0.0014 & 0.0063 & 1 \\
        
        XGB & KNN & 0.0049 & -0.0076 & -0.0014 & 1\\
        
        Light & GBM & 1.88E-7 & -0.0118 & -0.0061 & 1 \\
        
        Light & MLP & 2.16E-34 & 0.1605 & 0.1489 & 1 \\
        
        Light & RF & 9.43E-6 & 0.0068 & -0.0029 & 1 \\
        
        Light & KNN & 5.89E-17 & -0.0143 & -0.0104 & 1 \\
        
        GBM & MLP & 2.16E-40 & -0.1520 & -0.1395 & 1 \\
        
        GBM & RF & 0.0128 & 0.0009 & 0.0073 & 1 \\
        
        GBM & KNN & 0.0347  & -0.0067 & -0.0002 & 1 \\
        
        MLP & RF & 1.30E-36 & 0.1439 & 0.1558 & 1 \\
        
        MLP & KNN & 8.43E-36 & 0.1364 & 0.1486 & 1 \\
        
        RF & KNN & 4.92E-8 & 0.0051 & 0.0100 & 1 \\
    \noalign{\smallskip}\hline
    \end{tabular}
\end{table}
 
 \begin{table}[t]
    \centering
    \caption{$T$-test results for regression in Scenario 2.}
    \label{tab:ttest_reg_shadow}
    \begin{tabular}{llllll}
    \hline\noalign{\smallskip}
        ${G}_1$ & ${G}_2$ & $p$-value & Low. b. & Upp. b.  & \revmaj{${H}$} \\
        \noalign{\smallskip}\hline\noalign{\smallskip}
        XGB & Light & 1.99E-37 & 0.7565 & 0.8363 & 1 \\
        
        XGB & GBM & 3.00E-20 & -0.3563 & -0.2658 & 1 \\
        
        XGB & MLP & 1.13E-27 & -0.4830 & -0.3956 & 1 \\
        
        XGB & RF & 1.19E-6 & -0.1576 & -0.0731 & 1 \\
        
        XGB & KNN & 2.71E-6 & -0.1602 & -0.0710 & 1 \\
        
        Light & GBM & 1.95E-53 & -1.1401 & -1.0748 & 1 \\
        
        Light & MLP & 3.94E-60 & -1.2661 & -1.2053 & 1 \\
        
        Light & RF & 3.51E-57 & -0.9399 & -0.8836 & 1 \\
        
        Light & KNN & 1.42E-50 & -0.9437 & -0.8803 & 1 \\
        
        GBM & MLP & 3.50E-9 & -0.1656 & -0.0908 & 1 \\
        
        GBM & RF & 3.95E-16 & 0.1602 & 0.2314 & 1 \\
        
        GBM & KNN & 5.43E-15 & 0.1571 & 0.2338 & 1 \\
        
        MLP & RF & 3.89E-28 & 0.2904 & 0.3576 & 1 \\
        
        MLP & KNN & 2.75E-26 & 0.2871 & 0.3602 & 1 \\
        
        \textbf{RF} & \textbf{KNN} & \textbf{0.9867} & \textbf{-0.0344} & \textbf{-0.0350} & \textbf{0} \\
        \noalign{\smallskip}\hline
    \end{tabular}
\end{table}
 
Analyzing Tables~\ref{tab:ttest_reg_noshadow} and~\ref{tab:ttest_reg_shadow}, \revmaj{the hypothesis that ~\emph{XGBoost} and~\emph{GBM} are equivalents is rejected in Scenario~1.} 
The same applies to~\emph{RF} and~\emph{KNN} in Scenario~2. Therefore, it is clear that~\emph{LightGBM} is the one that best fits into the database obtained from this simulation campaigns, also presenting a diminished execution time, due to the fact that it has the smallest value for $MAE$ and the $t$-test confirms there is no other model with equivalent performance. 
It is worth to mention that~\emph{KNN} offers an acceptable performance while requiring an extremely low execution time, \revmaj{which suggests its use }
for similar applications with real-time regressions.

\section{Conclusions}
\label{sec:conclusions}

Since 3GPP \revmaj{HO} management relies basically on power \revmaj{level} comparisons, several inefficiencies \revmaj{arise} during such procedures. In this context, 
\revmaj{we presented a user-level simulation-based performance analysis of}
algorithms for classification and regression applications in 3GPP networks.
The classification \revmaj{aims to predict} the best \revmaj{HO} target, whereas the regression \revmaj{estimates the} download time and its completed percentage. 
Classical computational intelligence approaches, such as~\emph{KNN},~\emph{MLP},~\emph{SVM} and also recent fuzzy logic systems and latter gradient boosting machines were implemented.

The results indicate a valuable performance 
even in adverse propagation conditions while requiring short processing time. For classification, \emph{SVM} and~\emph{MLP} have the best performance, although the fuzzy system~\emph{SOFL} has similar accuracy with lower processing time. Regarding the regression applications,~\emph{LightGBM} is certainly the one that best adapts to these work conditions and presents minor mean absolute error and processing time. However, it is worth to mention that~\emph{KNN} offers extremely low time values.


\bibliographystyle{IEEEtran}
\bibliography{referencias.bib}
\end{document}